\documentclass[11pt]{article}

\usepackage[]{acl}

\usepackage{times}
\usepackage{latexsym}

\usepackage[T1]{fontenc}

\usepackage[utf8]{inputenc}

\usepackage{microtype}

\usepackage{inconsolata}

\usepackage{graphicx}

\usepackage{epsfig}
\usepackage{booktabs}

\usepackage[utf8]{inputenc} 
\usepackage[english]{babel}
\usepackage{array}
\usepackage{bm}
\usepackage{algorithm}
\usepackage{algorithmic}
\usepackage{multirow}
\usepackage{multicol}
\usepackage{graphicx}
\usepackage{amsmath} 
\usepackage{url} 
\usepackage{tikz}
\usetikzlibrary{arrows.meta, positioning}
\usepackage{subcaption}

\title{XL-DURel: Finetuning Sentence Transformers for Ordinal Word-in-Context Classification}

\author{Sachin Yadav, Dominik Schlechtweg\\
  Institute for Natural Language Processing, University of Stuttgart\\
  \texttt{sachindv42@gmail.com}, \texttt{schlecdk@ims.uni-stuttgart.de} \\}

\begin{document}
\maketitle
\begin{abstract}
We propose XL-DURel, a finetuned, multilingual Sentence Transformer model optimized for ordinal Word-in-Context classification. We test several loss functions for regression and ranking tasks managing to outperform previous models on ordinal and binary data with a ranking objective based on angular distance in complex space. We further show that binary WiC can be treated as a special case of ordinal WiC and that optimizing models for the general ordinal task improves performance on the more specific binary task. This paves the way for a unified treatment of WiC modeling across different task formulations.
\end{abstract}

\section{Introduction}
The Ordinal Graded Word-in-Context (OGWiC) task asks to predict the level of semantic proximity between two word usages on an ordinal scale \citep{schlechtweg2025comedi}. While it builds on the earlier WiC \citep{pilehvar2019wic} and GWiC \citep{armendariz-etal-2020-semeval} tasks, it can be distinguished from these by being formulated as an \textbf{ordinal classification} task \citep{sakai2021evaluating}. This is similar to ranking in that labels are inherently ranked, but also similar to classification in that exact labels have to be predicted for each test instance, instead of merely an ordering of instances as in ranking tasks. State-of-the-art OGWiC models employ pre-trained Language Models like XLM-R \citep{conneau2019unsupervised} to generate contextualized embeddings for the target word in two different contexts and finetune these with loss functions tailored for \textbf{binary} or \textbf{nominal} data such as the contrastive or the cross-entropy loss \citep[e.g.][]{Cassotti2023Xllexeme, Kuklin2025comediDeepchange}. We conjecture that these models do not sufficiently exploit the ranking signal provided by OGWiC training data.
In this study, we aim to overcome this limitation by employing loss functions directly optimizing ranking or regression objectives. We compare these against previous models trained with binary classification objectives and manage to outperform the latter on ordinal and binary data with a ranking objective based on angular distance in complex space. By improving performance on the binary and the ordinal formulation of the task through the same model, we pave the way for a unified treatment of WiC modeling. We publish our top-performing model, \textbf{XL-DURel}, which can be employed as highly optimized, fine-grained and multilingual contextualized embedder for word-meaning-related tasks.\footnote{\url{https://huggingface.co/sachinn1/xl-durel}}

\section{Related Work}
\subsection{WiC Task}
\label{sec:wic}

The challenge of capturing the dynamic semantics of words has led to the development of various evaluation benchmarks. One notable contribution in this area is the Word-in-Context (WiC) task and the corresponding dataset, introduced by \citet{pilehvar2019wic}. The WiC task is designed to assess context-sensitive word representations by framing it as a binary classification problem. In this task, each instance consists of a target word $w$, and two usages (or sentences) $u_1$ and $u_2$. The objective is to determine whether the meaning of the target word remains consistent across the two usages.
If the meaning is the same, the instance is labeled `TRUE' (or `1') as in pair (1,2):

\begin{itemize}
  \setlength\itemsep{0em}
    \label{:ex2}
    \item[(1)] The expanded \textbf{window} will give us time to catch the thieves.
    \item[(2)] You have a two-hour \textbf{window} of clear weather to finish working on the lawn.

\end{itemize}
If the meaning differs, it is labeled `FALSE' (or `0') as in pair (3,4):
\begin{itemize}
  \setlength\itemsep{0em}
    \label{:ex1}
    \item[(3)] There’s a lot of trash on the \textbf{bed} of the river.
    \item[(4)] I keep a glass of water next to my \textbf{bed} when I sleep.

\end{itemize}
Performance on this task is usually evaluated with Accuracy. The first WiC dataset by \citet{pilehvar2019wic} was constructed from sense-annotated lexical resources such as WordNet \citep{fellbaum2005wordnet}, VerbNet \citep{schuler} and Wiktionary\footnote{\url{https://www.wiktionary.org/}}. 

\subsection{GWiC Task}
The Graded Word Similarity in Context (GWiC) task \citep{armendariz-etal-2020-semeval}, introduced as part of SemEval-2020 Task 3, aims to evaluate how well computational models can capture graded word similarity in different contexts such as (5) and (6):
\begin{itemize}
  \setlength\itemsep{0em}
    \item[(5)] \label{item:exampleg1}
\ldots These young \textbf{men} displayed true Rajput chivalry. Akbar was so impressed with the bravery of these two \textbf{warriors} that he commissioned\ldots    
    \item[(6)] \label{item:exampleg2}
\ldots By night, she’s a top-ranking woman \textbf{warrior} in the Nine-Tailed Fox clan, charged with preserving the delicate balance between \textbf{man} and fox.
\end{itemize}
In one of the subtasks, participants were tasked with predicting the absolute similarity rating for each word pair within each context on a scale from 0 to 10. For word pair \textit{man} and \textit{warrior}, the gold similarity score is 7.88 in (5) and 3.27 in (6). The shared task used the Harmonic Mean of Pearson and Spearman correlations as an evaluation metric. It can thus be interpreted as a mixture of a regression and a ranking task.

\subsection{OGWiC Task}
\label{sec:ogwic}
The Ordinal Graded Word-in-Context (OGWiC) task was introduced as part of the CoMeDi shared task \citep{schlechtweg2025comedi}, focusing on nuanced and interpretable evaluation of word meaning in context. It aims to address the problems of the WiC and GWiC tasks by defining an ordinal classification task requiring participants to exactly reproduce the median annotated label for a word usage pair on a scale from 1 (unrelated) to 4 (closely related).\footnote{Find more details on the scale in Appendix \ref{sec:durel}.} For example, the pair (7,8) receives label 4 (identical) while pair (7,9) receives the lower label 2 (distantly related):
\begin{itemize}
  \setlength\itemsep{0em}
  \item[(7)] \ldots the dismissal last month of the commandant and two other generals of the provincial police, reportedly for \textbf{graft}.
  \item[(8)] We try to live with lies and corruption and fraud and \textbf{graft} and violence and exploitation and\ldots
  \item[(9)] The second, which is spread while warm on strips of coarse cotton, or strong paper, and wrapped directly about the \textbf{graft}, answering at once to tie and to protect it, is composed of equal parts of bees-wax, turpentine, and resin.
\end{itemize}
OGWiC is similar to the previous WiC and GWiC tasks, but limits the label set in predictions and penalizes stronger deviations from the true label. This makes OGWiC an \textbf{ordinal classification task} \citep[][]{sakai2021evaluating}, in contrast to binary classification (WiC) or ranking (GWiC). Predictions are evaluated against the median labels with the ordinal version of Krippendorff's $\alpha$ \citep[][]{krippendorff2018content}.

Two models excelled in the CoMeDi shared task \citep{Choppa2025comedi,Kuklin2025comediDeepchange}: \textbf{XL-LEXEME} \citep{Cassotti2023Xllexeme} builds upon the Sentence Transformers architecture \citep{reimers2019sbert} and employs a bi-encoder framework within a Siamese network. Vectors are initialized with XLM-RoBERTa \citep[XLM-R, ][]{conneau2019unsupervised} and their similarity is directly optimized using a contrastive loss function \citep{hadsell2006dimensionality}, which minimizes the distance between embeddings of sentences with the same meaning (label `TRUE') while maximizing the distance between embeddings of sentences with different meanings (label `FALSE') around a pre-selected margin. At test time, the model predicts the similarity between two usages using the finetuned base model and thresholds it to infer ordinal labels (see Section~\ref{sec:thres}). A similar approach is taken by the \textbf{DeepMistake} model \citep[][]{Arefyev2021Deep}. Vectors are initialized with XLM-R, sentences concatenated and jointly encoded. Then the target word vectors are extracted and jointly fed into a binary classification head. The model is finetuned using the cross-entropy loss. Similar to XL-LEXEME, DeepMistake is trained on binary WiC-like data. At test time, the model predicts the probability of label `TRUE' and thresholds it to infer ordinal labels.

\begin{table*}[t]
\small
\centering
\begin{tabular}{cccclll}
\toprule
\textbf{Dataset} & \textbf{Train} & \textbf{Dev} & \textbf{Test} & \textbf{Cosine} & \textbf{Binary} & \textbf{Ordinal} \\
\midrule
\multirow{4}{*}{CoMeDi} & \multirow{4}{*}{47,833} & \multirow{4}{*}{8,287} & \multirow{4}{*}{15,332} 
  & $4 \rightarrow 1.0$   & $4 \rightarrow 1$ & \\
& & & & $3 \rightarrow \frac{2}{3}$ & $3 \rightarrow 1$ &  \\
& & & & $2 \rightarrow \frac{1}{3}$ & $2 \rightarrow 0$ &  \\
& & & & $1 \rightarrow 0.0$ & $1 \rightarrow 0$ &  \\
\midrule
\multirow{2}{*}{WiC} & \multirow{2}{*}{251,972} & \multirow{2}{*}{8,381} & \multirow{2}{*}{6,400} 
  & $1 \rightarrow 1.0$ &  & $1 \rightarrow 4$ \\
& & & & $0 \rightarrow \frac{1}{3}$ & & $0 \rightarrow 2$ \\
\midrule
WiC+CoMeDi & 299,805 & 16,668 & 21,732  & as above & as above & as above \\
\bottomrule
\end{tabular}
\caption{Dataset statistics and label mappings.}
\label{tab:mapping-overview}
\end{table*}

\section{Data}
\label{sec:data}
\subsection{Ordinal WiC}
\label{ordinal-wic}
We use the OGWiC data provided by the CoMeDi shared task organizers \citep{schlechtweg2025comedi}, available in starting kit 1.\footnote{\url{https://comedinlp.github.io/}} The data comprises 71k word usage pairs sampled from ordinal WiC datasets across multiple languages, including Chinese, English, German, Norwegian, Russian, Spanish and Swedish. (See Table \ref{tab:data-dwug} in Appendix \ref{appendix:ogwic} for details.) 
The data was cleaned in various steps: Initially, instances with fewer than two annotations or those marked with any ``Cannot decide'' were excluded. Instances with significant annotator disagreement (more than one point on the scale) were also removed. A median judgment was calculated for each instance, retaining only integer medians for task consistency. The data was split by language, with 70\% allocated to training, 20\% to testing, and 10\% to development, ensuring that no target word overlapped between these splits. (See Table \ref{table:data_distribution} in Appendix \ref{appendix:ogwic} for details.)

\subsection{Binary WiC}
In addition to the CoMeDi data, our study incorporates the datasets used for training the XL-LEXEME model (see Section \ref{sec:ogwic}), which combines three established multilingual benchmarks: \texttt{XL-WiC} \citep{raganato-etal-2020-xl}, \texttt{MCL-WiC} \citep{martelli2021mclwic}, and \texttt{AM2iCo} \citep{liu-etal-2021-am2ico}. These benchmarks are widely used for evaluating word meaning in context.

\paragraph{XL-WiC} is a multilingual extension of the original WiC dataset \citep{pilehvar2019wic}, containing over 112k sentence pairs across 12 languages: Bulgarian, Chinese, Croatian, Danish, Dutch, Estonian, Farsi, French, German, Italian, Japanese, and Korean. Training data is available for German, French, and Italian while development and test sets are provided for all 12 languages.
Most of the data was automatically extracted from WordNet or Wiktionary sense inventories without direct human annotation. The dataset is distributed together with the original English WiC dataset comprising roughly 7K sentence pairs.

\paragraph{MCL-WiC} (Multilingual and Cross-lingual Word-in-Context Disambiguation) comprises approximately 10k sentence pairs spanning five languages: Arabic, Chinese, English, French, and Russian. The dataset contains data for two distinct subtasks: (i) multilingual WiC classification within individual languages, and (ii) cross-lingual classification comparing sentences from different languages. The dataset is specifically designed to evaluate model performance across both high- and medium-resource language settings. Unlike XL-WiC, which relies on sense inventories, MCL-WiC is entirely human-annotated.

\paragraph{AM2iCo} (Adversarial and Multilingual Meaning in Context) contains roughly 196k instances spanning 14 language pairs and 15 typologically diverse languages, including English, German, Russian, Japanese, Korean, Mandarin Chinese, Arabic, Indonesian, Finnish, Turkish, Basque, Georgian, Urdu, Bengali, and Kazakh. The dataset supports evaluation of word meaning in context both within individual languages and across different languages, with a particular focus on low-resource scenarios. AM2iCo is constructed by automatically extracting WiC pairs from Wikipedia, and then filtering them through human validation and adversarial filtering.\footnote{Adversarial filtering is a strategy to make a dataset harder and more useful by removing easy examples that models can solve without actually understanding the task.}\\

\noindent \citeauthor{Cassotti2023Xllexeme} constructed the training set for XL-LEXEME by merging the official training splits from the three above-described datasets. To further augment the training data, they randomly sampled 75\% of each dataset’s development data and added it to the training pool. The remaining 25\% of the development data was reserved for hyperparameter tuning and validation.

As part of our study, we concatenate the CoMeDi shared task dataset and the XL-LEXEME dataset into a unified resource. For clarity, we refer to the XL-LEXEME dataset concatenation henceforth as ``WiC (train/dev)''. We further refer to the CoMeDi shared task data as ``CoMeDi (train/dev/test)''. Additionally, we include the original WiC and MCL-WiC test datasets for evaluation in our experiments. We refer to these as ``WiC (test)''. Statistics for the final datasets are given in Table \ref{tab:mapping-overview}.\footnote{Unintentionally, we skipped 10 files in \citeauthor{Cassotti2023Xllexeme}'s training data package, i.e., AM2iCo dev ar-en/bn-en, XL-WiC dev bg-bg/da-da/en-en/et-et/fr-fr/zh-zh, WiC dev en-en, MCL-WiC dev en-en. In total, these are 12,512 instances.}

\subsection{Label Mapping}
\label{sec:label_mapping}
As summarized in Table \ref{tab:mapping-overview}, we apply a systematic label mapping procedure to align the datasets for unified model training and evaluation. Specifically, we transform binary and ordinal labels to cosine-like labels (interval $[0, 1]$) if needed for the respective loss function used for training (see Section \ref{sec:loss}). Similarly, we transform ordinal labels to binary labels if needed. As summarized in Table \ref{tab:mapping-overview}, for ordinal-to-cosine mapping, we utilize \texttt{Min-Max-Scaling} to map labels to the interval $[0, 1]$.
This maps the ordinal labels as follows: 1 $\rightarrow$ 0.0, 2 $\rightarrow \frac{1}{3}$, 3 $\rightarrow \frac{2}{3}$, and 4 $\rightarrow$ 1.0.  For binary-to-cosine mapping, we map label 1 (same sense) to cosine label 1.0 to align with annotation level 4 (identical) on the ordinal scale (cf. Table \ref{table:scale-durel} in Appendix \ref{sec:durel}). Binary label 0 (different sense) is mapped to cosine label $\frac{1}{3}$ to align with level 2 (polysemy) on the ordinal scale, based on the assumption that most pairs of usages, especially from the same target word, will be semantically related, e.g. by contiguity or similarity. For ordinal-to-binary mapping, we group ordinal labels 1 and 2 as binary label 0, and labels 3 and 4 as binary label 1, which is motivated by the idea that ordinal label 2 (polysemy) is a relation \textit{between} senses while ordinal label 3 (context variance) is a variation \textit{within} a sense (see Appendix \ref{sec:durel}). This mapping is needed in some cases for evaluation. Following the same logic as for binary-to-cosine, for binary-to-ordinal mapping we assign binary label 1 to ordinal label 4, and binary label 0 to ordinal label 2.

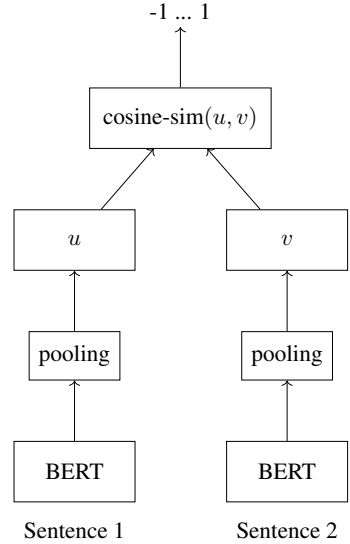
\begin{figure}[t]
\centering
\scalebox{0.8}{
\begin{tikzpicture}[
    node distance=1cm and 1.5cm,
    roundnode/.style={rectangle, draw, minimum width=2cm, minimum height=1cm, align=center},
    poolnode/.style={rectangle, draw, minimum width=1.5cm, minimum height=0.8cm, align=center},
    cosine/.style={rectangle, draw, minimum width=3cm, minimum height=1cm, align=center}
]

\node[roundnode] (bert1) {BERT};
\node[roundnode, right=of bert1] (bert2) {BERT};
\node[poolnode, above=of bert1] (pool1) {pooling};
\node[poolnode, above=of bert2] (pool2) {pooling};
\node[roundnode, above=of pool1] (u) {$u$};
\node[roundnode, above=of pool2] (v) {$v$};
\node[cosine, above=of u, xshift=1.75cm] (cosine) {cosine-sim$(u, v)$};
\node[above=of cosine] (similarity) {-1 ... 1};
\draw[->] (bert1) -- (pool1);
\draw[->] (bert2) -- (pool2);
\draw[->] (pool1) -- (u);
\draw[->] (pool2) -- (v);
\draw[->] (u) -- (cosine);
\draw[->] (v) -- (cosine);
\draw[->] (cosine) -- (similarity);

\node[below=0.2cm of bert1] {Sentence 1};
\node[below=0.2cm of bert2] {Sentence 2};
\end{tikzpicture}
}
\caption{SBERT architecture at inference.}
\label{sbert-architecture}
\end{figure}

\section{Model}
\label{sbert-architecture-text}
We employ Sentence-BERT~\citep[SBERT,][]{reimers2019sbert}, a modification of BERT designed to generate semantically meaningful sentence embeddings for efficient semantic similarity comparison. Unlike standard BERT, which requires joint processing of sentence pairs, SBERT uses a \textit{siamese} or \textit{triplet} network architecture. Each sentence is independently encoded by a BERT-based model with shared weights and pooled, resulting in fixed-size sentence embeddings.
SBERT is implemented using the Sentence Transformers library~\citep{reimers2019sbertmodule}, which offers a flexible interface for finetuning SBERT models with various loss functions. Most of these aim to adjust base model parameters so that similarities between sentence embeddings align with gold similarity values. A standard choice for the optimized similarity metric is the Cosine Similarity or its inverse, the Cosine Distance \citep{salton1986introduction}. See Figure~\ref{sbert-architecture} for an illustration of the SBERT architecture.

\subsection{Target Word Marking}
\label{sec:target_token_marking}
The WiC task requires semantic disambiguation at the token level rather than at the sentence level. This presents a challenge for SBERT, which is primarily designed for sentence-level embedding and comparison. We adopt XL-LEXEME's (see Section \ref{sec:ogwic}) strategy to adapt sentence embeddings to focus on specific target words within their contexts by marking the target word in each usage $u$ with special tokens:
\begin{align*}
u &= w_1, \ldots, \langle t \rangle, w_{t_i}, \ldots, w_{t_i + k}, \langle /t \rangle, \ldots, w_N
\end{align*}
where $\langle t \rangle$ and $\langle /t \rangle$ denote the opening and closing markers for the target word $w_t$, and $w_i$ represents individual words in the sentence. Inputs are truncated to a maximum sequence length of 128 tokens. After truncation, we additionally append the \texttt{[CLS]} and the \texttt{[SEP]} token before and after the input sequence, respectively.

\subsection{Loss Functions}
\label{sec:loss}
We experiment with the following loss functions to optimize model performance. If not stated differently, we use the cosine similarity/distance and the loss is calculated as mean per batch.

\paragraph{Contrastive Loss} expects two embeddings \((u, v)\) and a binary label \(y \in  \{0,1\}\) as inputs \citep{hadsell2006dimensionality}. It drives the similarity between positive pairs towards 1 and that between negative pairs to decrease to a
margin. In the Sentence Transformers library, the loss is defined as:

\vspace{-0.3cm}
{\small
\begin{equation*}
\mathcal{L}= \frac{1}{2}\bigg(y \cdot d(u, v)^2 + (1 - y) \cdot \max(0, m - d(u, v))^2\bigg)
\end{equation*}
\normalsize}
where
\begin{itemize}
  \item \(d(u, v)\) is the distance between the embeddings and
  \item \(m\) is the margin hyperparameter, which specifies the minimum required distance between dissimilar pairs.
\end{itemize}
Selecting the optimal margin for different datasets may be challenging \citep{cosentloss}. Also, relative distances with ordered label sets with more than two classes cannot be encoded making the loss ill-suited for ranking tasks.

\paragraph{Cosine Similarity Loss} expects two  embeddings \((u, v)\) and a continuous similarity label \(y \in [0, 1]\) as inputs. It is defined as the mean squared difference between the predicted similarities between embeddings and the ground truth label:
\begin{equation*}
\mathcal{L} = \left\| \cos(u, v) - y \right\|_2
\end{equation*}
where
\begin{itemize}
\item \( \left\| \cdot \right\|_2 \) is the L2 norm.
\end{itemize}
The mean squared error is a common loss function used in regression tasks. However, according to \citet{cosentloss}, it is
unsuitable for classification tasks because noise does not follow a normal distribution \citep[cf.][]{ciampiconi2024surveytaxonomylossfunctions}.

\paragraph{CoSENT Loss} expects two embeddings \((u, v)\) and a continuous similarity label \(y \in [0, 1]\) as inputs \citep{cosentloss}. It trains the embeddings so that the higher the similarity label between pairs, the higher the similarity of their embeddings:

\vspace{-0.3cm}
{\small
\begin{equation*}
\mathcal{L} = \log \bigg( 1 + \sum_{y(u,v)>y(k,l)} \exp(\lambda (s(k, l) - s(u, v)) ) \bigg)
\end{equation*}
\normalsize}
where 
\begin{itemize}
\item \(s(u, v)\) is the similarity between the embeddings,
\item \(y(u,v)>y(k,l)\) defines the set of embedding pairs \((k, l)\) for which the ground truth label \(y(k,l)\) is smaller than \(y(u,v)\) and
\item \(\lambda\) is a hyperparameter for amplification.\footnote{The Sentence Transformers library makes two specific implementation choices: (i) Pairs \((k,l)\) that do not meet the condition \(y(u,v)>y(k,l)\) are masked by subtracting a large constant (i.e., \(10^{12}\)) from their score difference, making their contribution negligible in the exponential term. (ii) For numerical stability, a zero is appended to the set of all cosine similarities in a batch where \(y(u,v)>y(k,l)\) to guarantee numerical stability in cases where the set \(y(u,v)>y(k,l)\) is empty.}
\end{itemize}
The loss is computed as sum over all pairs \((u, v)\) in the batch. In contrast to the contrastive loss operating \textit{within} the sentence pairs, CoSENT focuses on maintaining ranking consistency \textit{between} the learned similarity of sentence
pairs within the entire set and their similarity labels \citep{cosentloss}. This also distinguishes it from the Cosine Similarity Loss which operates only on individual pairs and only implicitly optimizes ranking consistency.

\paragraph{AnglE Loss} expects two  embeddings \((u, v)\) and a continuous similarity label \(y \in [0, 1]\) as inputs. It uses the CoSENT Loss function (see above) with a different similarity measure, i.e., the angle difference in complex space \citep{li2023angleoptimized}. The AnglE Loss was introduced to address a key limitation of the cosine function: The gradient of the cosine function tends to approach zero as it nears its maximum or minimum values, which can hinder the optimization process. According to \citeauthor{li2023angleoptimized}, this is not the case for the angle difference in complex space.

\subsection{Optimization}
\label{optimization}
During training, the parameters of the base model are adjusted in order to minimize the respective loss function from Section \ref{sec:loss}. For all experiments, we keep the following settings constant: We use XLM-R-large as our base model and optimize with \texttt{AdamW}. We set the learning rate to $1\times10^{-5}$, the batch size to $32$ and use no weight decay ($0.0$). All other settings are kept at their default values. For Contrastive Loss we set the margin \(m=0.5\), and for the ranking losses (CoSENT, AnglE Loss) we use the default \(\lambda=20\). We train for 10 epochs, with a linear warm-up over 10\% of the total training steps. We evaluate at every 25\% of an epoch via Average Precision\footnote{We also tried Spearman correlation and did not observe a considerable difference in results.} (Contrastive Loss) or Spearman correlation (rest) between cosine similarities and gold labels on dev data (see Section \ref{subsec:evaluation-metrics}).\footnote{We also experimented with using the angle difference on dev for AnglE Loss, but did not outperform the cosine similarity.} The final checkpoint is chosen by highest performance on dev data.

The base model, XLM-R-large, contains 561M parameters. All experiments are conducted on a Linux-based server running Fedora 42, equipped with NVIDIA RTX A6000 GPUs (48 GB VRAM per GPU) and dual Intel Xeon CPUs. We utilize a single GPU per run and estimate the computational runtime per model run to be approximately 40–50 GPU hours.\footnote{GitHub Copilot was used to assist with coding during the implementation.}

\subsection{Thresholding}
\label{sec:thres}

For all models, we adopt the CoMeDi shared task baseline approach to map cosine similarities to ordinal labels. At test time, similarities are mapped to ordinal labels using three thresholds $\theta$, which are optimized on the dev set by minimizing the following loss function \citep[cf.][]{Choppa2025comedi}:
\[\mathcal{L} = 1 - \alpha(\mathbf{y}, \hat{\mathbf{y}}_{\theta})\]
where 
\begin{itemize}
    \item $\mathbf{y}$ are gold labels, 
    \item $\hat{\mathbf{y}}_{\theta}$ are predicted labels according to thresholds $\theta$ on similarity predictions $\hat{\mathbf{y}}$ and 
    \item $\alpha$ is Krippendorff's $\alpha$. 
\end{itemize}
Krippendorff’s $\alpha$ is the task evaluation metric (see Section \ref{subsec:evaluation-metrics}). We aim to find optimal values for $\theta$.
This threshold optimization is performed on the dev data and separately for each language to account for language-specific distributional differences in similarity scores. It uses the Nelder--Mead simplex method \citep{nelder1965simplex}. (Find induced thresholds in Appendix \ref{sec:thresholds}.)

\subsection{Baseline Models}
\label{sec:baselines}
In our experiments, we employ a number of simple baseline models, as described below. All models use thresholding as explained in Section \ref{sec:thres} for mapping similarities to ordinal labels.

\paragraph{SBERT} uses the cosine similarity on a non-finetuned SBERT model initialized with XLM-R-Large.

\paragraph{XL-LEXEME} uses XLM-R-Large as base model and was finetuned with SBERT using the Contrastive Loss on WiC train and dev (see Sections \ref{sec:ogwic} and \ref{sec:data}).

\paragraph{XL-LEXEME CoMeDi} is the XL-LEXEME result reported in the CoMeDi shared task. Notably, it achieved the second-best performance. We use this as a baseline in our evaluation.
        
\paragraph{DeepMistake CoMeDi} is the DeepMistake result reported in the CoMeDi shared task. It achieved the best performance. It uses XLM-R-Large as base model and was finetuned with the cross-entropy loss for binary classification on MCL-WiC train and dev, the Spanish subset of XLWSD \citep{Pasini2021XLWSD} and a binarized version of the Spanish DWUG dataset \citep{Zamora2022lscd} (see Sections \ref{sec:ogwic} and \ref{sec:data}). Because Spanish DWUG is part of our test data, we report additional average performance without Spanish in Section \ref{sec:experiments}.

\begin{figure*}[t]
    \centering \includegraphics[width=0.8\textwidth]{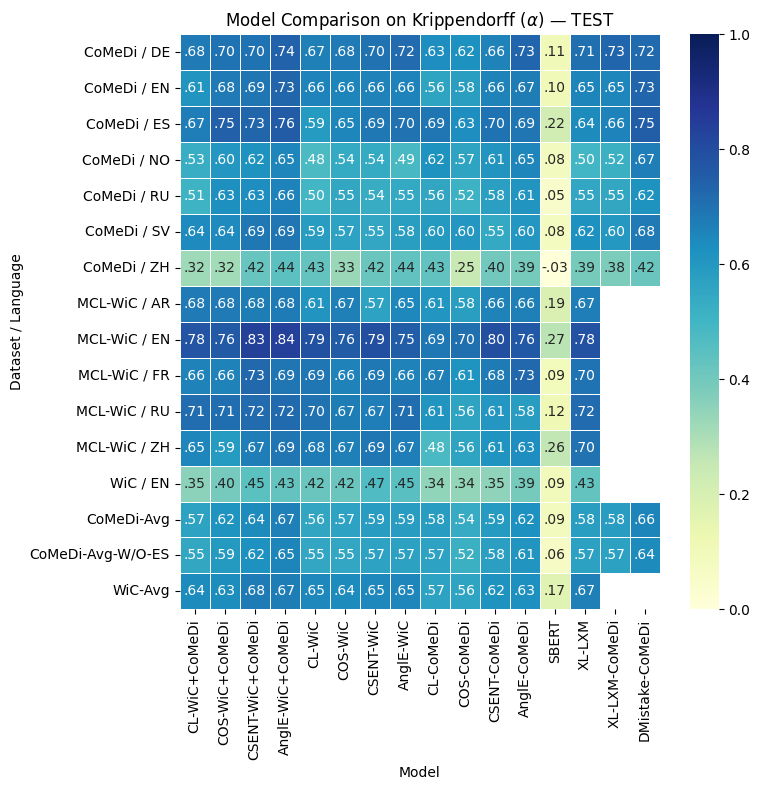}
    \caption{Model evaluation with Krippendorff's $\alpha$ on binary and ordinal test data. CL = Contrastive Loss, COS = Cosine Similarity Loss, CSENT = CoSENT Loss, AnglE = AnglE Loss, XL-LXM = XL-LEXEME, DMistake = DeepMistake.}
    \label{fig:loss-krip}
\end{figure*}

\section{Evaluation}
\label{subsec:evaluation-metrics}

Following the CoMeDi shared task, we use ordinal \textbf{Krippendorff's \(\alpha\)} \citep{krippendorff2018content} as evaluation measure for ordinal classification.
It penalizes stronger deviations from the gold label more heavily. It has the additional advantage of controlling for expected disagreement and has been demonstrated to be superior to other measures such as Mean Absolute Error for ordinal classification \citep{sakai2021evaluating}. We also use Spearman’s rank correlation coefficient ($\rho$) between continuous similarities and gold ordinal labels to assess the rank alignment of model predictions. This enables us to evaluate performance without inducing thresholds. We further apply the nominal version of Krippendorff and the Spearman correlation for binary label evaluation. During training, similarities are compared to ordinal labels with Spearman correlation and to binary labels with the Average Precision.

\section{Experiments}
\label{sec:experiments}

We now test which loss functions (see Section \ref{sec:loss}) and data combinations (see Section \ref{sec:data}) improve performance on the ordinal CoMeDi test data over the baselines described in Section \ref{sec:baselines}. We additionally report performance on binary WiC data to understand whether optimization for ordinal data hurts the binary task. As finetuning is computationally expensive, we perform one run for each model.\footnote{We selectively re-ran models and observed variation of average performances between $\pm.01$--$.03$.}
Models are evaluated with Krippendorff's $\alpha$ as described in Section \ref{subsec:evaluation-metrics} based on their cosine predictions binned to ordinal labels (see Section \ref{sec:thres}).\footnote{We also experimented with using the angle difference on test for AnglE Loss, but did not outperform the cosine similarity.} All experiments follow the training setup described in Section~\ref{optimization}. Results are shown in Figure \ref{fig:loss-krip}.\footnote{In Appendix \ref{sec:spearman}, we give an additional performance evaluation on the raw cosine predictions with Spearman rank correlation.} 

\paragraph{Loss function} First of all, we see that finetuning has a strong effect on performance. For this, compare all models against SBERT, which is the only non-finetuned model. With a performance of .67, AnglE Loss achieves the best result on CoMeDi test data (`CoMeDi-Avg') when using both WiC and CoMeDi training data (AnglE-WiC+CoMeDi). With performances of .64 and .62 respectively, it is followed by the CoSENT Loss (CSENT-WiC+CoMeDi) and the Cosine Similarity Loss (COS-WiC+CoMeDi). All these models clearly outperform the published version of XL-LEXEME (XL-LXM), its retrained model version (CL-WiC) and the retrained version with additional ordinal data (CL-WiC+CoMeDi), which have performances of .58, .56 and .57, respectively. The top model AnglE-WiC+CoMeDi outperforms the latter ones on all languages and on average by a large margin. The same holds for the published XL-LEXEME result from the CoMeDi shared task (XL-LXM-CoMeDi), which is outperformed with .67 vs. .58. It further slightly outperforms the shared task winning model DeepMistake (.67 vs. .66). This is also the case if we exclude Spanish (`CoMeDi-Avg-W/O-ES'), which was reported additionally by the task organizers because DeepMistake was trained on part of the test data for Spanish. Notably, we reach this performance by optimizing one unified model while the DeepMistake result was obtained by optimizing multiple models tailored to specific languages.\footnote{We provide the finetuned AnglE-WiC+CoMeDi model under the name ``XL-DURel'' at 
\url{https://huggingface.co/sachinn1/xl-durel}. Find the code for reproducing our results at \url{https://github.com/sachinn12/XL-DURel}.}
On binary WiC (`WiC-Avg'), top performance is reached by the CoSENT Loss model relying on both binary and ordinal training data (CSENT-WiC+CoMeDi) with .68, closely followed by the AnglE Loss relying on the same data (AnglE-WiC+CoMeDi) and XL-LEXEME with .67, respectively. Hence, on the ordinal and the binary task ranking losses show top performance, where the advantage to the classification loss is more pronounced for the ordinal task. While the regression loss is not competitive for either task, it shows a clear advantage over the classification loss on the ordinal task. These results are in line with the motivations given for the loss functions in Section \ref{sec:loss}: AnglE and CoSENT Loss are explicitly optimizing a ranking objective, which exploits the inherent ordering of ordinal labels. Further, the AnglE Loss improves optimization over the CoSENT Loss and other cosine-based losses, presumably because it avoids killed gradients occurring with the cosine similarity.

\paragraph{Training data} Note that training on purely ordinal data yields good baseline performance across tasks, especially with the AnglE Loss (AnglE-CoMeDi). For binary data, this is also the case, but there is a larger performance difference to top models on the ordinal task (e.g. AnglE-WiC). Moreover, we clearly observe that combining ordinal and binary data improves performance on the ordinal task across all loss functions. Compare for example performances of AnglE-WiC/CoMeDi vs. AnglE-WiC+CoMeDi or CSENT-WiC/CoMeDi vs. CSENT-WiC+CoMeDi. There is a clear average performance improvement on the ordinal task. Similarly for performance on the binary task, but only for the ranking losses AnglE and CoSENT. However, for the Contrastive and the Cosine Similarity Loss the performance drop is negligible.

\paragraph{Ordinality} Regardless of training data, the ordinal training signal turns out to be beneficial for both tasks. In order to see this, compare e.g. CL-CoMeDi vs. AnglE-CoMeDi. The former binarizes the data while the latter keeps the ordinal information. The performance difference is striking with .58 vs. .62 for the ordinal task and, interestingly, .57 vs. .63 for the binary task. This indicates that fine-grained semantic proximity information helps the model to better learn binary meaning distinctions, which is also supported by the fact that the best purely ordinal model approaches the performance of the best purely binary model on the binary task (AnglE-CoMeDi vs. CL-WiC) despite being trained on much less and out-of-distribution data.

\section{Conclusion}
We compared several loss functions for classification, regression and ranking to finetune OGWiC models. Our top model outperformed previous models on ordinal and binary WiC data with a ranking objective based on angular distance in complex space. Overall, we found that using the AnglE Loss can be recommended, both for the ordinal and the binary WiC task. Similarly, mixing ordinal and binary training data turned out to be beneficial for both tasks. These results suggest that binary WiC can be treated as a special case of ordinal WiC and that optimizing models for the general ordinal task improves performance on the more specific binary task. In the future, WiC task setups should try to unify these approaches in order to make use of the full power of WiC training signals from multiple types of data. Further, we should try to optimize models more directly for ordinal classification instead of ranking. Currently, our model first predicts a dense similarity which we then discretize in an independent step to ordinal labels through thresholds. However, there are also loss functions directly optimizing for ordinal labels, like Cumulative Link models \citep{Vargas2020cumulativeordinal}. We would like to test such models for OGWiC. Moreover, similar to \citet{Loke2025comedi}, we would like to test Large Language Models such as Llama \citep{touvron2023llama}. These profit from massive amounts of parameters and training data and can be directly instructed to predict an ordinal number. Note, however, that our current approach has certain advantages over this: It is theoretically motivated by employing direct ranking optimization. Also, it is small and efficient making the model applicable to large amounts of data.

\section*{Limitations}

We tested our hypotheses using particular data, base models and training architectures. In future research, these should be varied to test whether they have an influence on effects. Specifically, it should be tested whether the AnglE Loss turns out to be beneficial for the ordinal and binary task with additional test data. It is also unclear why the AnglE Loss performs better with the cosine similarity than with the angle difference in complex space at test time although the latter is optimized during training.

\section*{Acknowledgments}
This study is an extension of Sachin Yadav's master thesis \citep{Yadav2025thesis}. We thank Lucas Möller for feedback regarding the implementation. We further thank Pierluigi Cassotti, Roksana Goworek and Haim Dubossarsky for help on reproducing XL-LEXEME results.

\bibliography{references,Bibliography-general,bibliography-self,bibliography-supervision-self}

\clearpage
\appendix

\section{Annotation Scale}
\label{sec:durel}
Unlike WiC, which is designed as a binary classification task, OGWiC employs an \textbf{ordinal classification} approach by assigning labels on a \textbf{four-point scale}. This four-point scale in Table~\ref{table:scale-durel}, follows the \textbf{DURel annotation framework} proposed by \citet{Schlechtwegetal18} which is based on Blank’s concept of semantic proximity \citep{Blank97XVI}. Unlike GWiC, labels are not transformed post-hoc and each level of the DURel scale has an exact linguistic interpretation as presented in Table~\ref{table:scale-durel}, where polysemy is located between identity, context variance, and homonymy \citep{Schlechtweg2023measurement}.

\begin{table}[t]
\parbox{.45\linewidth}{
\centering
\tabcolsep=0.11cm
\begin{tabular}{ll}
\multirow{4}{*}{$\Bigg\uparrow$} &4: Identical\\
 &3: Closely Related\\
 &2: Distantly Related\\
 &1: Unrelated\\
\end{tabular}
\label{tab:scale2}}
\hfill
\parbox{.45\linewidth}{
\centering
\tabcolsep=0.11cm
\begin{tabular}{ll}
\multirow{4}{*}{$\Bigg\uparrow$}&Identity\\
&Context Variance\\
&Polysemy\\
&Homonymy
\end{tabular}
\label{tab:blank}}
\caption{The DURel relatedness scale \citep[][]{Schlechtwegetal18} on the left and its interpretation from \citet[][p. 33]{Schlechtweg2023measurement} on the right.}\label{table:scale-durel}
%\vspace{-1.2cm}
\end{table}

According to \citet[][p. 22--23]{Schlechtweg2023measurement}, the pair (1,2) below is classified as identical as the referents of two uses of the word \textbf{arm} are both prototypical representatives of the same extensional category corresponding to the concept ‘a human body part’:
\begin{enumerate}
    \item[(1)] \label{item:example1}
    [...]\ and taking a knife from her pocket, she opened a vein in her little \textbf{arm}, [...]    
    \item[(2)] \label{item:example2}
    [...]\ and though he saw her within reach of his \textbf{arm}, [...]
\end{enumerate}
The use pair (1,3) is classified as context variance as both referents still belong to the same extensional category, but one is a non-prototypical representative. Hence, there is some variation in meaning, e.g. the arm of a statue loses the function of the physical arm to be lifted:
\begin{enumerate}
    \item[(3)] \label{item:example3}
        [...]\ when the disembodied \textbf{arm} of the Statue of Liberty jets spectacularly out of the sandy beach. [...]
\end{enumerate}
The use pair (1,4) would be classified as polysemy as the two referents of arm belong to different extensional categories, but the corresponding concepts still hold a semantic relation (in this case a similarity relation regarding physical form).
\begin{enumerate}
 \item[(4)] \label{item:example4}
        \ It stood behind a high brick wall, its back windows overlooking an \textbf{arm} of the sea [...]
\end{enumerate}
In contrast, the referents of arm in the homonymic pair (1,5) belong to different extensional categories and the corresponding concepts do not hold a semantic relation:
\begin{enumerate}
\item[(5)] \label{item:example5}
        And those who remained at home had been heavily taxed to pay for the \textbf{arms}, ammunition; fortifications, [...]
\end{enumerate}

\section{CoMeDi Data}
\label{appendix:ogwic}

Table \ref{tab:data-dwug} shows the source datasets used for the CoMeDi shared task. Table \ref{table:data_distribution} shows the number of data instances per language and split for the OGWiC subtask after cleaning.

\begin{table}[t]
\centering
\scriptsize
\tabcolsep=.05cm
\begin{tabular}{l l l l l l l}
\toprule
          \textbf{Dataset} & \textbf{LG} & \textbf{Reference} & \textbf{JUD} & \textbf{VER} & \textbf{KRI} & \textbf{SPR} \\\midrule 
     ChiWUG & ZH & \citet{Chen2023chiwug} & 61k & 1.0.0 & .60 & .69\\
    \midrule
     DWUG & EN & \citet{Schlechtweg2021dwug} & 69K & 3.0.0 & .63&.55\\
     DWUG Res. & EN & \citet[][]{Schlechtweg2024dwugs} & 7K& 1.0.0 & .56 & .59 \\
    \midrule
     DWUG & DE & \citet{Schlechtweg2021dwug} & 63K & 3.0.0 & .67 & .61  \\
     DWUG Res. & DE &\citet[][]{Schlechtweg2024dwugs} & 10K&  1.0.0 & .59& .7 \\
     DiscoWUG & DE & \citet{Kurtyigit2021discovery} & 28K & 2.0.0 & .59 &.57\\
     RefWUG & DE & \citet{Schlechtweg2023measurement} & 4k & 1.1.0 & .67 & .7\\
     DURel & DE & \citet{Schlechtwegetal18} & 6k & 3.0.0 & .54 & .59\\
     SURel & DE & \citet{haettySurel-2019} & 5k & 3.0.0 & .83 & .84\\
    \midrule
     NorDiaChange & NO & \citet{kutuzov2022nordiachange} & 19k & 1.0.0 & .71 & .74\\
    \midrule
     RuSemShift & RU & \citet{rodina2020rusemshift} & 8k & 1.0.0 & .52 & .53 \\
     RuShiftEval & RU & \citet{rushifteval2021} & 30k & 1.0.0 & .56 & .55\\
     RuDSI & RU & \citet{Aksenova2022rudsi} & 6k & 1.0.0 & .41 & .56\\
    \midrule
     DWUG & ES & \citet{Zamora2022lscd} & 62k & 4.0.1 & .53 & .57\\ 
%    & DiaWUG & ES & \citet{Baldissin2022diawug} & 8k\\
    \midrule
     DWUG & SV & \citet{Schlechtweg2021dwug} & 55K & 3.0.0 & .67& .62\\
     DWUG Res. & SV & \citet[][]{Schlechtweg2024dwugs} & 16K& 1.0.0 & .56& .65\\
    \bottomrule
\end{tabular}
\caption{Datasets used for the CoMeDi shared task. All are annotated on the DURel scale. Spearman and Krippendorff values for RuShiftEval are calculated as average across all time bins. `LG' = Language; `JUD' = Number of judgments; `VER' = Dataset version; `KRI' = Krippendorff's $\alpha$; `SPR' = Weighted mean of pairwise Spearman correlations; `Res.' = Resampled.}
\label{tab:data-dwug}
\end{table}

\begin{table}[t]
\centering
\begin{tabular}{cccc}
\toprule
\textbf{Language}  & \textbf{Train} & \textbf{Dev} & \textbf{Test} \\ \midrule
German   & 8,279  & 1,663  & 3,141  \\ 
English  & 5,910  & 863  & 2,444  \\ 
Swedish  & 5,457 & 871  & 1,245   \\ 
Chinese & 10,833   & 2,532  & 3,240  \\
Spanish & 4,821   & 621  & 1,497  \\
Russian & 8,029  & 1,126  & 2,285  \\ 
Norwegian& 4,504  & 611  & 1380  \\ \midrule
\textbf{Total} & \textbf{47,833}  & \textbf{8,287} &\textbf{15,332} \\ \bottomrule
\end{tabular}
\caption{Number of data instances per language and split for the OGWiC subtask after cleaning.}
\label{table:data_distribution}
\end{table}

\section{Thresholds}
\label{sec:thresholds}

Find the thresholds for mapping cosine similarity to ordinal labels for AnglE-WiC+CoMeDi and XL-LEXEME in Table \ref{tab:thresholds}. These were induced on the dev data as described in Section \ref{sec:thres}.

\begin{table*}[!hp]
\centering
\begin{tabular}{cc|ccc|ccc}
\toprule
\textbf{Dataset} & \textbf{Language} & \multicolumn{3}{c|}{\textbf{AnglE-WiC+CoMeDi}} & \multicolumn{3}{c}{\textbf{XL-LEXEME}} \\
\midrule
CoMeDi & ZH   & .577 & .677 & .793 & .495 & .650 & .655 \\
CoMeDi & EN   & .325 & .483 & .612 & .418 & .607 & .682 \\
CoMeDi & DE   & .330 & .465 & .600 & .339 & .565 & .651 \\
CoMeDi & NO   & .210 & .339 & .488 & .390 & .414 & .522 \\
CoMeDi & RU   & .255 & .491 & .615 & .249 & .511 & .749 \\
CoMeDi & ES   & .297 & .521 & .628 & .212 & .455 & .788 \\
CoMeDi & SV   & .290 & .452 & .564 & .419 & .646 & .672 \\
MCL-WiC & AR  &  ~~~~~~~~~~     & .634 &       &       & .741 &       \\
MCL-WiC & ZH  &       & .766 &       &       & .814 &       \\
MCL-WiC & EN  &       & .668 &       &       & .705 &       \\
MCL-WiC & FR  &       & .623 &       &       & .667 &       \\
MCL-WiC & RU  &       & .594 &       &       & .775 &       \\
WiC     & EN  &       & .551 &       &       & .752 &       \\
\bottomrule
\end{tabular}
\caption{Thresholds for AnglE-WiC+CoMeDi and XL-LEXEME.}
\label{tab:thresholds}
\end{table*}

\section{Spearman Results}
\label{sec:spearman}
\begin{figure*}[!hp]
    \centering \includegraphics[width=0.8\textwidth]{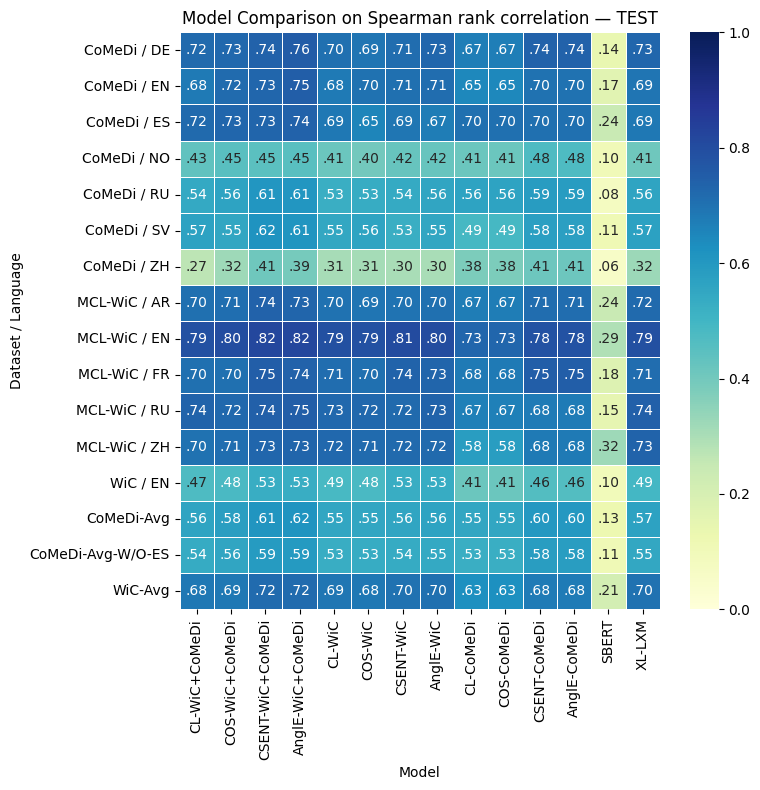}
    \caption{Model evaluation with Spearman's $\rho$ on binary and ordinal test data. CL = Contrastive Loss, COS = Cosine Similarity Loss, CSENT = CoSENT Loss, AnglE = AnglE Loss, XL-LXM = XL-LEXEME.}
    \label{fig:loss-spear}
\end{figure*}

Figure \ref{fig:loss-spear} shows model performances measured with Spearman correlation and no thresholding.

\end{document}